\documentclass{article}

\usepackage[preprint]{neurips_2026}
\workshoptitle{AI for Meta-Science Workshop}

\usepackage[utf8]{inputenc}
\usepackage[T1]{fontenc}
\usepackage{hyperref}
\usepackage{url}
\usepackage{booktabs}
\usepackage{amsfonts}
\usepackage{amsmath}
\usepackage{nicefrac}
\usepackage{microtype}
\usepackage{xcolor}
\usepackage{graphicx}
\usepackage{float}
\usepackage{longtable}
\usepackage{array}
\usepackage{ragged2e}
\newcolumntype{P}[1]{>{\RaggedRight\arraybackslash}p{#1}}

\title{Kurate: Scalable Scientific Quality Analysis}

\author{
Matthew J. Vowels \\
Kivira Health, U.S.A.\\
\texttt{matt@kivira.health}
\And
Jamie Cummins \\
University of Bern, Bern, Switzerland
}

\newcommand{\elmcite}{\citep{spiritconsortelm}}
\newcommand{\CorpusN}{4,347}
\newcommand{\RCTN}{3,913}
\newcommand{\QualityDims}{8}

\begin{document}

\maketitle

\begin{abstract}
Scientific search systems can find papers that are relevant to a question, but they generally do not assess the quality of the evidence that those papers provide. We present Kurate, a system that uses large language models (LLMs) to assess the quality of published studies. Kurate uses both the paper and its related documents (e.g., the study's trial registration and protocol), and links each of its judgments to the passage of text on which that judgment is based. We applied Kurate to a corpus of \CorpusN{} papers (\RCTN{} of which report randomized trials) and scored each paper on \QualityDims{} dimensions of study design and reporting: specifically, statistical power, causal identification, preregistration, selective reporting, measurement validity, analysis prespecification, reporting transparency, and conflict of interest and funding. Across the corpus, we found that papers most often exhibited issues with statistical power, selective reporting, and analysis prespecification, although average quality differed between clinical areas. When compared against expert annotations of 60 held-out clinical-trial documents, the information Kurate extracted matched the expert label in 221/242 protocol scorepoints and 294/370 results-publication scorepoints, with AC1 $0.94$ and $0.81$, respectively. Using a well-reputed, high quality clinical trial as a worked example, we show how a single paper's overall grade breaks down into separate judgments, with each linked to specific evidence from the trial's registration, protocol, and published report. Together, these results show that large-scale quality assessment of this kind is feasible, and that it can be used to address meta-scientific research questions.
\end{abstract}

\section{Introduction}
Scientific literature search is fundamental to science generation, evidence synthesis, and evidence-based decision-making. In doing this search for
papers which match a given topic, intervention, disease, or claim, researchers traditionally operate from the assumption that peer review has already filtered out low-quality work, and that the all entries in the corpus of literature carry high-quality evidentiary signals. However, modern research (in particular, the field of meta-science) recognises that this is not necessarily the case: a paper can be peer-reviewed and directly relevant to the question being asked, yet still be unsuitable for use as primary evidence for many potential reasons. For instance, the study results may be derived from selective reporting/outcome switching, uncorrected multiplicity, numerical errors in reporting, or inappropriate controls or randomisation (among many other possibilities). Manual systematic reviews and GRADE certainty assessments
(GRADE being a standard framework for rating how confident one can be in a body
of evidence) were developed to handle exactly this distinction
\citep{cochranehandbook2023,grade2008,borah2017analysis}. 

The aspects of a study which such appraisal should target are informed by a set
of recurring problems in the clinical literature. These include incomplete
registration and reporting, differences between the outcomes specified before a
trial began and the outcomes reported in the published paper, and selective
publication of results. They also include trials whose small samples, or whose
reliance on surrogate endpoints (i.e., outcomes such as reductions in brain amyloid, which are used as proxies for slower cognitive decline in Alzheimer's disease), can make the
evidence appear stronger than it is
\citep{anderson2015compliance,mathieu2009comparison,turner2008selective,dechartres2013sample,prasad2015surrogate,alexander2021aducanumab}.

Here we present Kurate, a software pipeline which automates this appraisal step. Using an architecture consisting of both deterministic checks and agentic large language models, Kurate ingests, processes, and assesses papers against a set of research quality criteria derived from contemporary methodological and meta-scientific standards, with judgements tied directly to specific related sections of the manuscript or research materials. Importantly, the resulting structured record for each paper is retained and may be used for several purposes, including individual paper evaluation, study quality-augmented literature searches, and broader cross-literature evaluation across clinical domains,
interventions, or funders. We first describe each stage of the Kurate pipeline.
We then provide a small-scale evaluation of Kurate's information extraction against expert
annotations of published trials. We demonstrate how Kurate can be used for scientometric evaluation using a corpus of \CorpusN{} medical and
pharmacological papers. Lastly, we provide an example of how Kurate's scoring works for individual trials, and how these scores can be traced back to the documents and judgments which produced them \citep{sprint2015}.

\section{Related Work}
\label{sec:relatedwork}

The limitations of human time are already evident in how often relevant study materials go unchecked. For instance, there is growing evidence that the majority of study registrations are not checked during peer review \citep{syed2025,compare2019}. Existing automated tools have therefore been developed to address the time-intensiveness of these activities. Several existing automated research-integrity tools each focus on one specific part of the review process. RobotReviewer assesses the risk of bias in individual trial reports and extracts the text which supports each assessment \citep{robotreviewer2016}. Trialstreamer continuously identifies newly published randomized trials and extracts structured information from them, such as PICO elements (population, intervention, comparison, and outcome), sample sizes, and risk-of-bias signals \citep{trialstreamer2020}. RegCheck compares study registrations against the corresponding published papers in a structured way \citep{regcheck2026}.

A second line of work focuses on rule-based checks of whether a paper contains values which are mathematically impossible or internally inconsistent. statcheck examines whether reported test statistics are consistent with the reported $p$-values \citep{nuijten2016statcheck}. The GRIM test examines whether reported means are arithmetically possible given the sample size \citep{brown2017grim}, and GRIMMER-style checks apply the same logic to standard deviations and related summary statistics \citep{jung2025scrutiny}. Other packages, such as Metacheck, aim to combine multiple automated checks of research outputs into a single framework to which new checks can be added \citep{metacheck2026}.

Other scalable tools, such as ReviewerZero \citep{reviewerzero2026} or SciScore \citep{menke2020sciscore}, cover a broader range of checks, but still treat the single paper as the unit of evaluation. More recently, LLM-based approaches have also been developed for evaluating research integrity. INSPECT-AI, for example, is an LLM-assisted tool built around the INSPECT-SR framework for assessing the trustworthiness of randomized trials \citep{inspect-ai,inspect-sr}. As such, nearly all existing tools evaluate individual papers. Some, like RegCheck or statcheck, can be applied at scale but each covers only one type of check \citep{regcheck2026,nuijten2016statcheck}. Tools built to support systematic reviews, such as RobotReviewer \citep{robotreviewer2016}, work on the studies included in one review rather than on a field as a whole. In contrast, Kurate collects each paper together with its related documents (e.g., registrations), extracts information from these artefacts, scores the paper along several quality dimensions, and evaluates whether the paper is appropriate for guiding evidence-based decision-making, thereby supporting meta-scientific analysis across the whole corpus.

\section{Kurate}
\label{sec:kurate}
Kurate processes each paper in four stages: (1) collecting the paper and its
related documents, (2) extracting information which requires interpretation of
the text, (3) scoring the paper's quality based on pre-defined quality dimension, and (4) deciding whether the paper
should be retained as supporting evidence when results are combined across
studies (e.g., in a meta-analysis or more general review of the literature). We refer to the last
of these as the evidence-use decision. The first stage uses only fixed rules.
From the published report, Kurate extracts its bibliographic information, full text
(divided into short passages), tables, figures, authors, Medical Subject
Headings (MeSH; the standard keywords used to index biomedical articles),
funding statements, and registry identifiers, along with a record of any
problems encountered during this parsing. Kurate then adds whatever related documents are
available for that paper including its registry entry, protocol, funding
records, citations, and integrity signals (e.g., retractions or corrections),
using a combination of fixed rules and external tools (e.g., ClinicalTrials.gov for trial registrations, PubPeer for post-publication commentary, and PubMed APIs for indexing metadata and report content). Each associated study document is stored separately,
so that the information in each can later be compared. Importantly, these steps do not yet
involve any judgment about the study itself, as their purpose is only to collect
the available evidence (see Fig.~\ref{fig:pipeline} and Appendix~\ref{sec:rubric}).

Some information cannot be extracted with fixed rules, because identifying it
requires interpretation of the scientific text. This information is instead
extracted by a set of LLM-based agents (i.e., LLM-based programs which each carry
out a specific extraction task). The extraction tasks performed by these agents include the study's identifying
labels, its trial-registration details, outcome designation (e.g., primary or
secondary outcomes), what each treatment group received, evidence on sample size and
statistical power, the flow of participants through the study (e.g., from a CONSORT flow diagram), disclosures of
funding and conflicts of interest, and comparisons of registration/protocol-paper reporting consistency. Because many of the evaluations for these extractions require interpretation of text,  we require any fact which requires such interpretation to either be extracted by an agent which links it to a specific
passage in the text which evidences this extraction, or else be marked as unavailable.

Kurate's full set of quality evaluation dimensions is more extensive than those
used in demonstration in this manuscript (see Appendix~\ref{sec:rubric}). For the current corpus, Kurate scores \QualityDims{} dimensions, namely: statistical power,
causal identification (i.e., whether the study design supports a causal
conclusion, for example through randomization), preregistration, selective
reporting, measurement validity, analysis prespecification, reporting
transparency, and conflict of interest and funding. Each paper receives a score
on each dimension, where a positive score indicates that the available evidence
supports the paper's quality on that dimension and a negative score indicates
that there may be an issue associated with this dimension. From these scores, Kurate calculates a normalized score
between 0 and 1, a grade from A to F, and an evidence-use decision (which either consists of an include or exclude judgement, or an explicit flag for manual review). 

\begin{figure}[H]
    \centering
    \includegraphics[width=\linewidth]{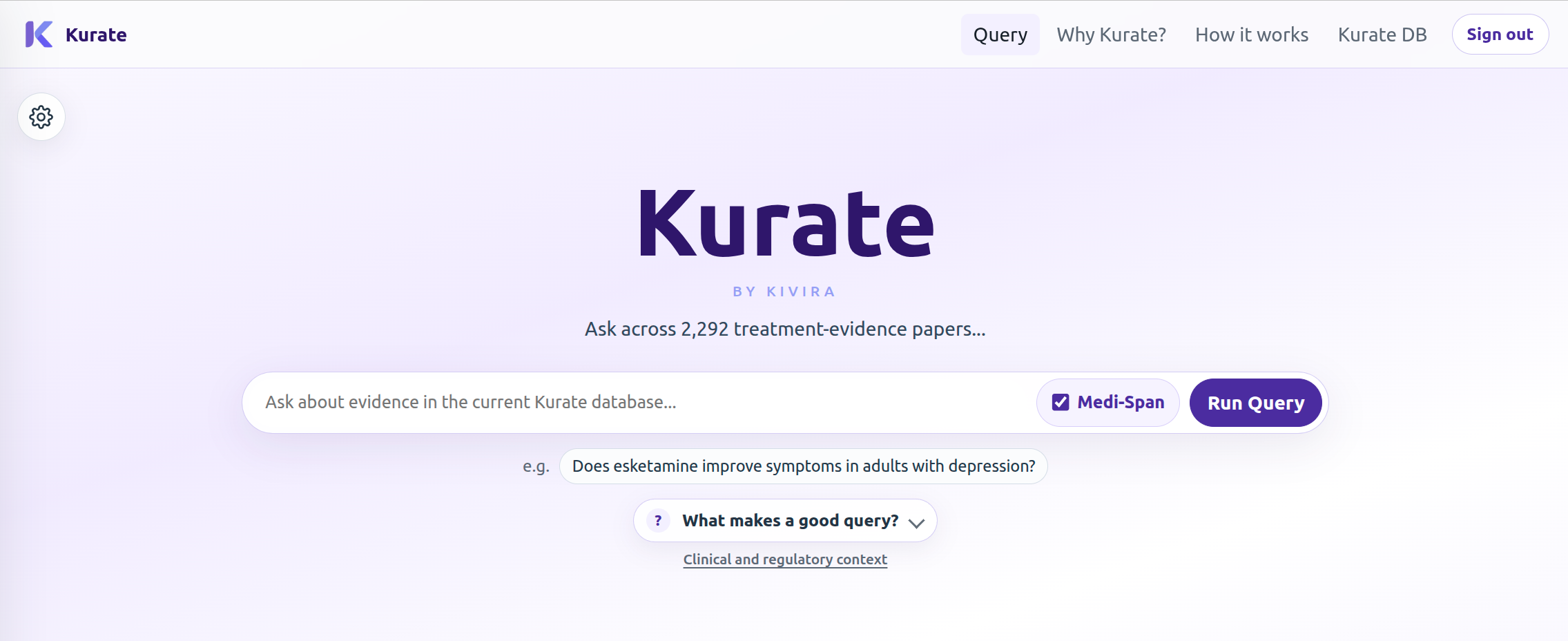}
    \caption{The Kurate search interface. }
    \label{fig:interfacea}
\end{figure}

\begin{figure}[H]
    \centering
    \includegraphics[width=1.1\linewidth]{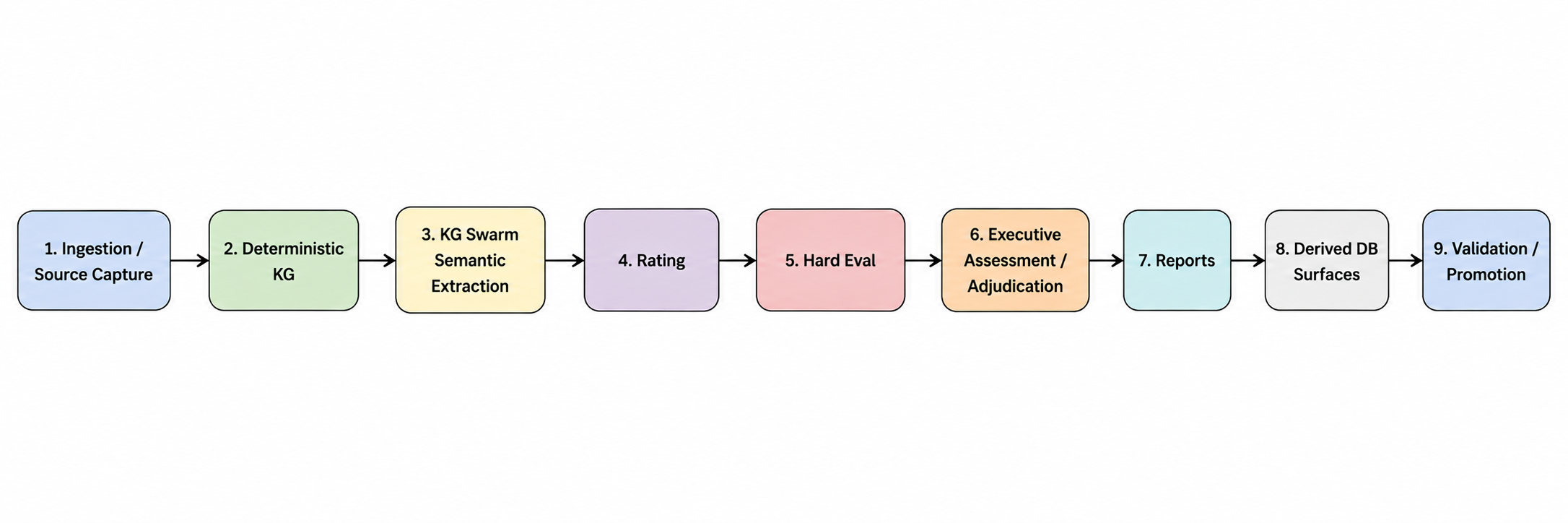}
    \caption[End-to-end Kurate pipeline.]{
    The Kurate pipeline. Stage names in bold correspond to the boxes in the figure.
    \textbf{Ingestion/source capture} collects the published report of a study together with its linked documents, and extracts the bibliographic information, full text, tables, figures, registry and protocol identifiers, funding statements, and a record of any problems encountered during parsing.
    \textbf{Deterministic Knowledge Graph} uses fixed rules (i.e., without LLMs) to build a knowledge graph (KG), which is a database of linked records connecting papers, related documents, authors, funders, registrations, text passages, tables, and figures.
    \textbf{KG swarm semantic extraction} uses a set of LLM-based agents to extract information which requires interpretation of the text. This includes which outcomes were primary or secondary, what the treatment and comparison conditions consisted of, which disorders were studied, evidence on sample size and statistical power, the flow of participants through the study, whether the paper agrees with its registration and protocol, and information on funding and conflicts of interest.
    \textbf{Rating} scores this information against Kurate's quality dimensions (Appendix~\ref{sec:rubric}). This produces a score for each dimension (positive where the evidence supports the study's quality, and negative where it indicates a problem), an overall score scaled to lie between 0 and 1 (the normalized score), and a letter grade.
    \textbf{Hard eval} checks for serious problems which can rule out the use of a paper as evidence regardless of its grade. Examples of such issues can include a document which is not an empirical report, the absence of randomization, conflicts between the registration and the published report, large amounts of missing data, and primary claims which the results do not support.
    \textbf{Executive assessment/adjudication} combines the scores, the results of these checks, and any uncertainties or caveats into a final decision about whether the paper should be used as evidence.
    \textbf{Reports} produces a report for each paper which shows the evidence, the document each piece of evidence came from, the scores, the caveats, and the reasons for the final decision.
    \textbf{Derived Database surfaces} updates the summary tables used for searching and for corpus-level analysis (e.g., summaries by topic, medication, and funder).
    \textbf{Validation/promotion} checks each paper's record for completeness and consistency, and adds fully processed papers to the curated database used for browsing, searching, and meta-scientific analysis.
    }
    \label{fig:pipeline}
\end{figure}

In addition, Kurate consists of a set of "deal-breaker" criteria which, if not met, lead immediately to a decision to not use the evaluated study as evidence. These checks cover cases including compromised
randomisation, undisclosed problematic registration-paper reporting conflicts, unsupported primary claims, and other serious integrity signals. All
of these outputs are stored in a SQLite database, together with the extracted
text passages, tables, figures, quality summaries, and
indexes of papers by medication, topic, and funder. Importantly, knowledge-graph records are also generated for each paper based on these extractions and metadata.

\begin{figure}[H]
    \centering
    \includegraphics[width=0.7\linewidth]{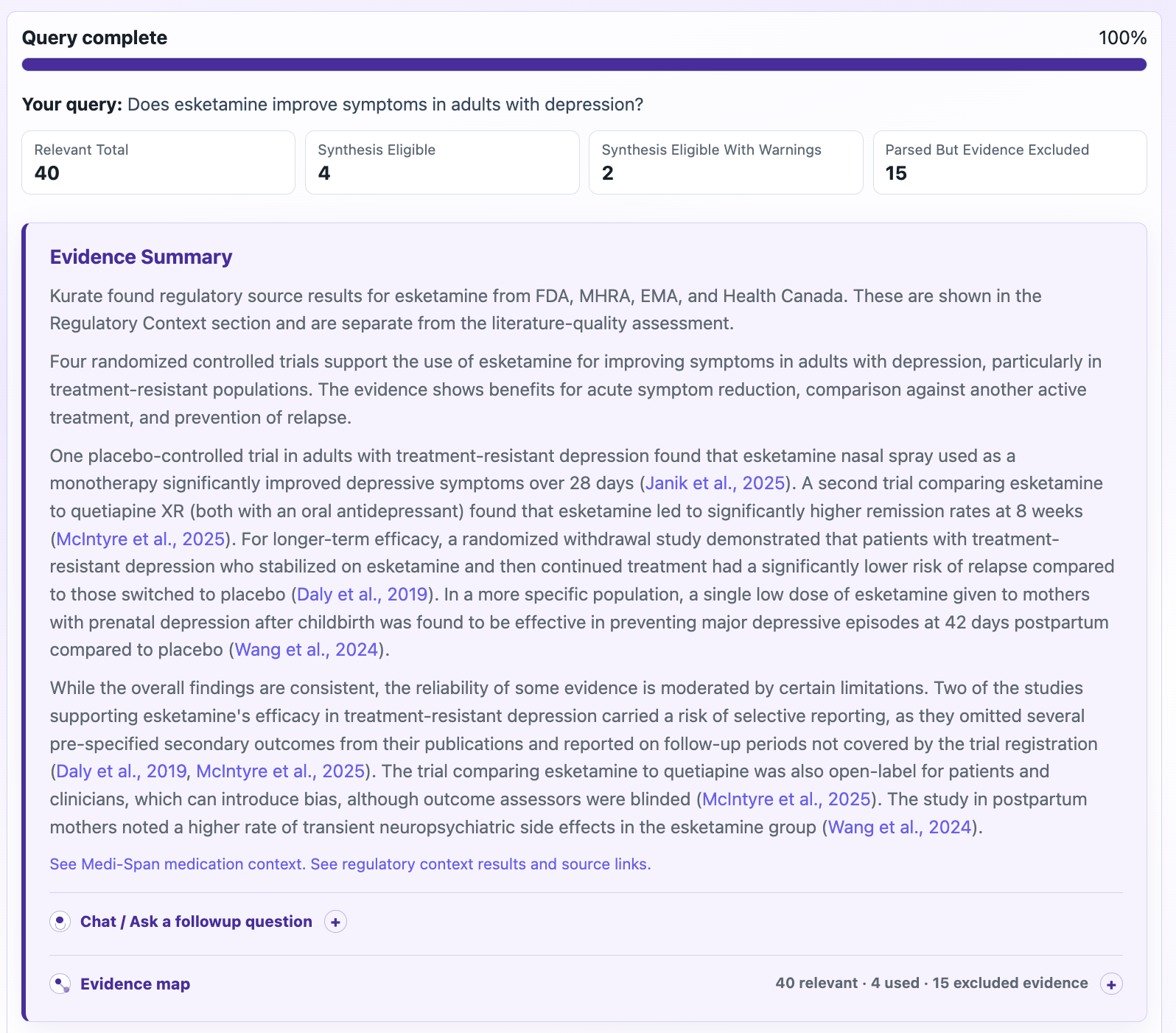}
    \caption{The evidence summary which Kurate returns for an example query about esketamine.}
    \label{fig:interfaceb}
\end{figure}

Finally, Kurate provides a graphical interface for searching the generated
database of evaluated papers. The interface encourages users to phrase their queries in PICO format
(i.e., specifying the population, intervention, comparison, and outcome of
interest), and returns a summary of the evidence across the studies which match
the query (see Figures~\ref{fig:interfacea} and~\ref{fig:interfaceb} for an
example query about esketamine). Importantly, this interface facilitates users' searches of the literature (and its related research quality) using natural language querying.

\section{Evaluating Kurate}
\label{sec:benchmark}

Every corpus-level result in Section~\ref{sec:results} depends on Kurate
extracting information from each paper correctly. To test this directly, we
compared Kurate's extraction against SPIRIT-CONSORT-ELM \elmcite, an
independent benchmark dataset of 200 clinical-trial documents (100 trial
protocols, and the 100 publications which report those trials' results). In this
benchmark, human experts marked, for each item of the standard reporting
checklists for trial protocols (SPIRIT) and trial results (CONSORT), the exact
passage of text in each document which reports that item.

\begin{figure}[t]
    \centering
    \includegraphics[width=0.8\linewidth]{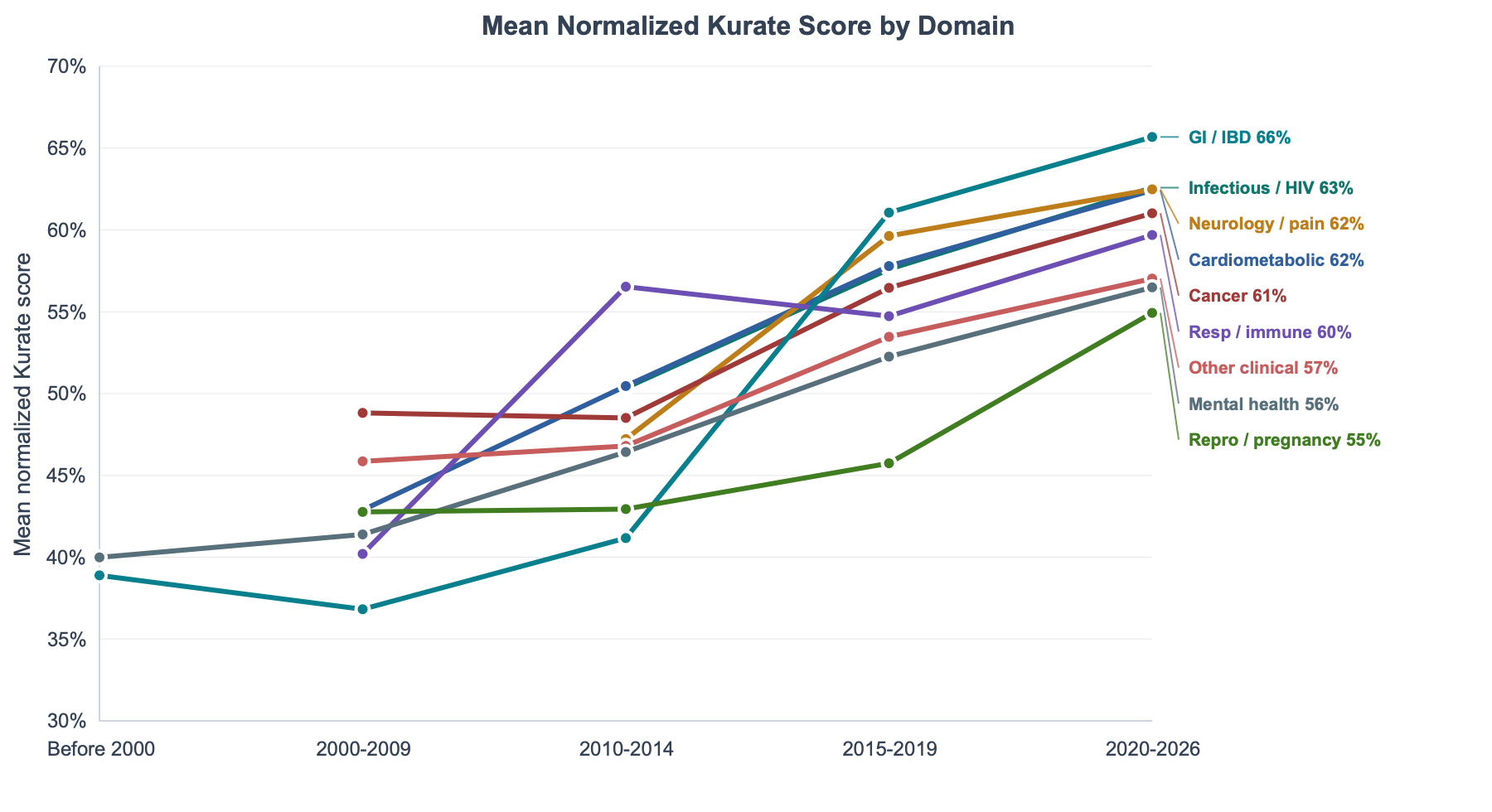}
    \caption{Mean normalized Kurate scores by publication era and clinical domain. Points are shown where a domain has at least five papers in that era; endpoint labels identify trajectories and latest observed-era scores.}
    \label{fig:domain-trajectories}
\end{figure}

This benchmark is informative for three reasons. First, this represents a comparison against human labellled text, which enables evaluation of Kurate relative to independent human performance, rather than simply evaluating Kurate's output on its own terms. Second, the benchmark reports how
well individual human annotators perform on the same task, which indicates how
high a score can realistically be expected: that is, an individual expert annotator
agrees with the final consensus labels at AC1 $0.89$ (AC1 being a
chance-corrected measure of agreement, described below), rather than at $1.0$. Accounting for such reasonable scholarly disagreement ensures a fairer and more realistic evaluation of the LLM-based tool. 
Third, the benchmark's authors report the performance of their own automated
system, in which a retrieval model (PubMedBERT) finds candidate passages and a
language model (GPT-5) then answers the reporting questions. This provides a
published point of comparison, and we refer to this system as the
\emph{baseline} throughout. Critically, this can tell us whether Kurate's more complex architecture and pipeline provides marginal improvements compared to a simpler (albeit, still complex) pipeline. 

In the main text, we report whether the single value which Kurate passes on to
its later stages (i.e., scoring, the evidence-use decision, and reporting)
matches the expert label. In Appendix~\ref{sec:benchmark-appendix}, we also
report a more lenient measure (whether the correct value appeared anywhere among
the candidate values Kurate extracted before selecting one), together with the
full methods. All results are based on documents which were not used during
Kurate's development (i.e., the benchmark's own test and validation sets), and
each document type was scored once.

\begin{table}[t]
\centering
\small
\setlength{\tabcolsep}{5pt}
\caption{Accuracy of Kurate's information extraction, compared with expert annotators and a published automated system. \textbf{Upper block:} for each reporting item, the number of held-out documents in which the value used by Kurate matched the expert label, with the proportion in parentheses. ``---'' indicates that the item is not annotated for that type of document (results publications do not state who was planned to be eligible, and protocols contain no findings to report). \textbf{Lower block:} Gwet's AC1 \citep{gwet2008} measures agreement after correcting for the agreement expected by chance. It is the measure reported by the benchmark's authors, and is therefore the only measure for which human and baseline figures are available. The 95\% CI row gives Kurate's confidence interval for mean AC1. The benchmark reports the human and baseline figures only as overall means across both document types, so each figure appears in both columns. They are also means over all 119 benchmark questions, whereas Kurate's figures are means over only the 9 (protocols) and 15 (results publications) questions for which Kurate extracts information. See Appendix~\ref{sec:benchmark-appendix}.}
\label{tab:benchmark}
\begin{tabular}{@{}>{\raggedright\arraybackslash}p{6.1cm}cc@{}}
\toprule
\textbf{Information Kurate extracts} & \textbf{Trial protocols} & \textbf{Results publications} \\
\midrule
\multicolumn{3}{@{}l}{\textit{Study design and population}} \\
\quad Who was eligible to take part & 30/30 (1.00) & --- \\
\quad What the intervention was & 30/30 (1.00) & 29/30 (0.97) \\
\quad What it was compared against & 30/30 (1.00) & 25/30 (0.83) \\
\quad How many participants & 30/30 (1.00) & 38/46 (0.83) \\
\quad How the outcome was defined & 24/30 (0.80) & 26/29 (0.90) \\
\addlinespace
\multicolumn{3}{@{}l}{\textit{Statistical planning}} \\
\quad How the analysis would be done & 29/29 (1.00) & --- \\
\quad How the sample size was justified & 26/30 (0.87) & 17/30 (0.57) \\
\addlinespace
\multicolumn{3}{@{}l}{\textit{Reported findings}} \\
\quad $p$-values & --- & 21/22 (0.95) \\
\quad Effect sizes & --- & 37/46 (0.80) \\
\quad Summary statistics & --- & 35/46 (0.76) \\
\quad Confidence intervals & --- & 19/30 (0.63) \\
\quad Percentages and proportions & --- & 10/16 (0.62) \\
\addlinespace
\multicolumn{3}{@{}l}{\textit{Reporting practice}} \\
\quad Whether data are available & 5/5 (1.00) & 15/18 (0.83) \\
\quad Who funded the study & 17/28 (0.61) & 22/27 (0.81) \\
\midrule
\textbf{Total} & \textbf{221/242 (0.91)} & \textbf{294/370 (0.79)} \\
\midrule
\multicolumn{3}{@{}l}{\textit{Chance-corrected agreement, AC1 (and F1), over the same held-out documents}} \\
\quad \textbf{Kurate} & \textbf{0.94} (F1 0.93) & \textbf{0.81} (F1 0.88) \\
\quad 95\% CI for Kurate AC1 & [0.870, 1.000] & [0.725, 0.900] \\
\quad Human expert annotator & 0.89 & 0.89 \\
\quad Published automated baseline & 0.80 (F1 0.82) & 0.80 (F1 0.82) \\
\bottomrule
\end{tabular}
\end{table}

Table~\ref{tab:benchmark} shows the results for each item, along with the
comparison against the human and baseline figures. On trial protocols, Kurate
agrees with the expert labels at AC1 $0.94$, which is notably higher than the published
baseline ($0.80$). Importantly, the performance of an individual human expert ($0.89$) lies
within the 95\% confidence interval of Kurate's estimate ($0.870$ to $1.000$;
Appendix~\ref{sec:benchmark-appendix}). On results publications, Kurate reaches
AC1 $0.81$, which is approximately equal to the baseline and lower than the
human figure. On F1 (a measure which combines the proportion of extracted values
that are correct with the proportion of correct values that are extracted),
Kurate scores higher than the baseline for both document types. However,
because Kurate's figures cover only the subset of benchmark questions for which
it extracts information, these comparisons are not like-for-like and may favor
Kurate (Appendix~\ref{sec:benchmark-appendix}).

The item-level results show where the difference between the two document
types comes from. From protocols, Kurate correctly extracts the information
describing what a trial planned to do (who was eligible, which intervention was
given, what it was compared against, how many participants were planned, and how
the data would be analyzed) in every held-out document. At the same time, the same item can be much harder to extract from one type of document than
from the other. Kurate locates the sample-size justification in $87\%$ of
protocols, but in only $57\%$ of results publications. This is because a power
calculation is carried out before a study begins: it is expected to appear in a
protocol, whereas a results publication usually mentions it only briefly, if at
all. 

In contrast to the high levels of performance described above, a clear limitation of the current Kurate workflow relates to extraction of numeric information. Kurate's accuracy was lowest for such detailed numerical results,
particularly for confidence intervals and reported percentages (both approximately
$0.62$). Importantly, part of this lower accuracy reflects ambiguity in the benchmark
itself: $88$ of the $120$ numerical passages in the held-out results
publications contain more than one distinct number, so the correct answer is
not always a single, well-defined value. However, an ideal automated system conducting such extraction would disambiguate such cases more effectively, and this represents an area for improvement for Kurate.

\begin{figure}[H]
    \centering
    \includegraphics[width=\linewidth]{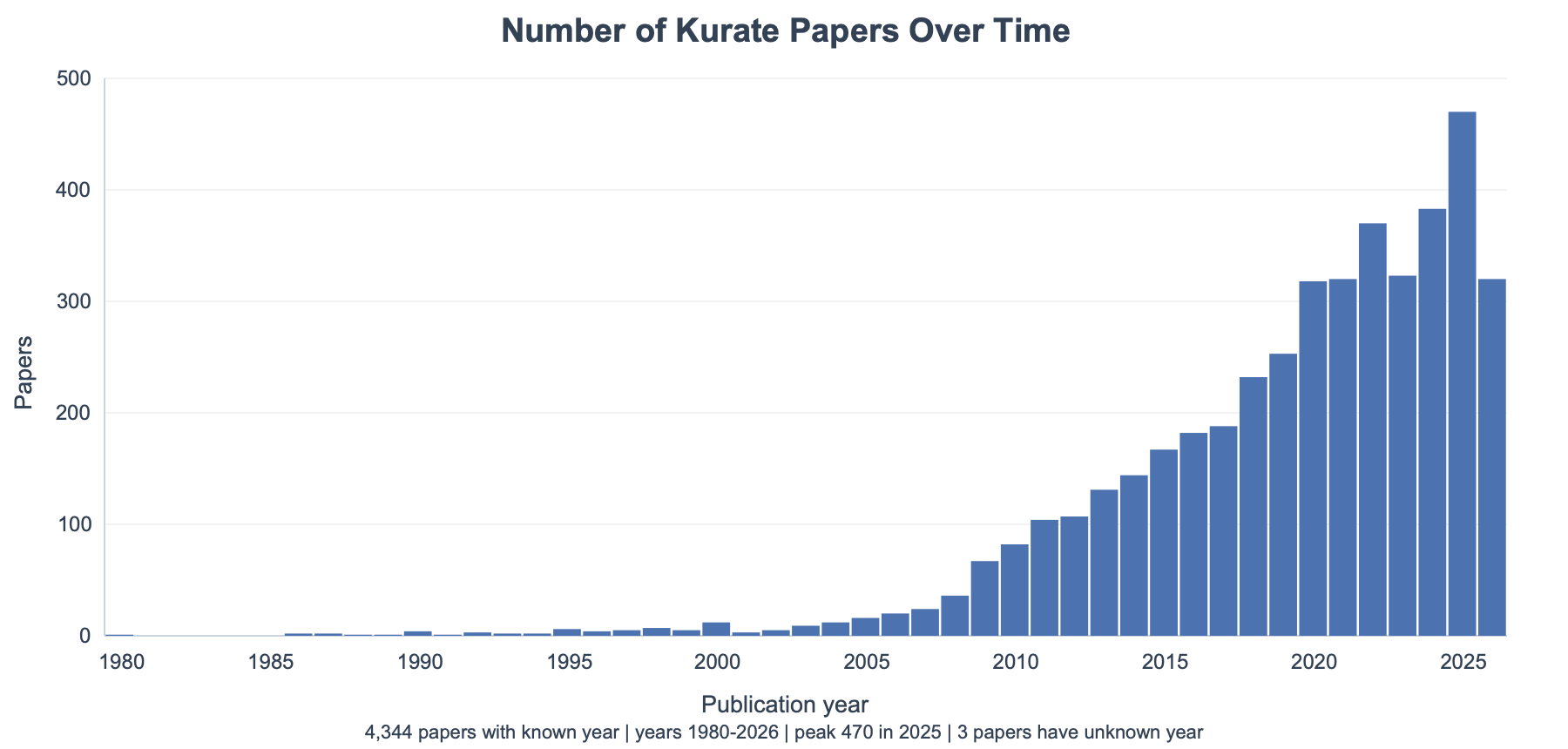}
    \caption{Number of papers by publication year in the current Kurate corpus among papers with a known publication year (4,344 of \CorpusN{} papers).}
    \label{fig:year-histogram}
\end{figure}

\section{Literature Quality Evaluation Using Kurate}
\label{sec:results}

In the above section, we described a small-scale evaluation of Kurate's extraction performance. Given the overall high levels of performance (with the exception of numeric extractions), we next applied Kurate to evaluate research quality at scale. 

\subsection{Corpus Selection}
\label{sec:corpus-targeting}
We selected the current corpus based on open-access availability of clinical and pharmacological papers. Consequently, this corpus does not represent a fully random sample of the biomedical literature. Wherever possible, each record was based on an empirical report article in PMC. The early stages of corpus construction focused on drug treatments for major depression; later stages expanded to randomized trials of commonly-used medications, covering cardiometabolic, respiratory/allergy/immunology, infectious disease/HIV, cancer, pain, arthritis/osteoporosis, gastrointestinal/IBD, reproductive/pregnancy, and mental-health topics. Candidate papers were identified using PubMed/PMC identifiers, MeSH and free-text terms for disorders and interventions, and a limited search of the related-article links which PubMed lists for a set of starting papers. Editorials, comments, news items, biographies, other documents which are not empirical reports, and records which failed Kurate's completeness checks were excluded. Registry records, protocols, statistical analysis plans (SAPs), and other related documents were collected only after the empirical report had been selected. The arXiv snapshot used for the results in this work is from a specific artefact labelled ``4,347 papers | 2026-09-10 09:17 UTC''; the corresponding query-term index metadata records \texttt{generated\_at} as \texttt{2026-09-10T09:17:19.314287+00:00}. This snapshot contains 604,183 text passages, 12,142 parsed tables, 13,347 parsed figures, 11,245 medication-index rows, and 2,095,352 knowledge-graph metadata rows.

\subsection{Quality Results}
\label{sec:corpus-results}
The Kurate quality grades assigned to papers varied significantly: 280 papers received an A, 934 a B, 1,425 a C, 1,471 a D,
236 an E, and 1 an F. In total, 27.9\% of papers were graded A or B, and
60.7\% were graded A, B, or C. Notably, Kurate's evidence-use decisions were stricter
than its grades: 225 papers were marked ``include'', 1,835 ``flag'', and 2,287
``exclude''. An excluded paper remains in the corpus, but is not used as
supporting evidence when results are combined across studies in Kurate responses to user queries. A flagged paper
may be used as supporting evidence, but with caveats or a recommendation for
manual checking. If both include and flag decisions are counted as retaining the
paper, 47.4\% of papers are retained, but only 17.6\% of all papers are both
retained and graded A or B. Table~\ref{tab:grade-evidence-use} cross-tabulates
quality grade against the final evidence-use decision.

\begin{table}[t]
\centering
\small
\setlength{\tabcolsep}{5pt}
\caption{Kurate quality grade by final evidence-use decision in the targeted \CorpusN{}-paper corpus. Counts are papers. Retained is the sum of \texttt{include} and \texttt{flag}; \texttt{exclude} papers remain in the corpus but are not used as supporting evidence in downstream synthesis.}
\label{tab:grade-evidence-use}
\begin{tabular}{lrrrrr}
\toprule
Grade & Include & Flag & Retained & Exclude & Total \\
\midrule
A & 71 & 137 & 208 & 72 & 280 \\
B & 100 & 456 & 556 & 378 & 934 \\
C & 44 & 606 & 650 & 775 & 1,425 \\
D & 10 & 570 & 580 & 891 & 1,471 \\
E & 0 & 65 & 65 & 171 & 236 \\
F & 0 & 1 & 1 & 0 & 1 \\
\midrule
Total & 225 & 1,835 & 2,060 & 2,287 & \CorpusN{} \\
\bottomrule
\end{tabular}
\end{table}

\begin{table}[t]
\centering
\small
\setlength{\tabcolsep}{4pt}
\caption{Quality results for each clinical domain in the targeted \CorpusN{}-paper corpus, sorted from highest to lowest mean normalized score (range 0--1). Only domains with at least 25 papers are shown. A/B/C is the percentage of papers graded A, B, or C; Excluded is the percentage of papers for which Kurate's evidence-use decision was ``exclude''. All percentages describe Kurate's outputs, not independently verified properties of the studies.}
\label{tab:corpus}
\begin{tabular}{p{3.2cm}rccc}
\toprule
Domain & $N$ & Mean score & A/B/C & Excluded \\
\midrule
Overall corpus & \CorpusN{} & 0.564 & 60.7\% & 52.6\% \\
Infectious disease / HIV & 349 & 0.602 & 72.5\% & 51.3\% \\
Neurology / pain & 138 & 0.593 & 63.0\% & 44.9\% \\
Cardiometabolic & 1,122 & 0.592 & 68.2\% & 51.2\% \\
Cancer & 843 & 0.580 & 65.5\% & 63.0\% \\
Resp. / allergy / immune & 152 & 0.569 & 60.5\% & 52.0\% \\
GI / IBD & 119 & 0.542 & 48.7\% & 40.3\% \\
Other clinical & 344 & 0.539 & 55.8\% & 53.2\% \\
Mental health & 1,185 & 0.526 & 51.0\% & 49.2\% \\
Repro. / pregnancy & 95 & 0.485 & 37.9\% & 50.5\% \\
\bottomrule
\end{tabular}
\end{table}

The breakdown in Table~\ref{tab:exclusion-breakdown} highlights in particular the 2,287 papers with a final
evidence-use decision of \texttt{exclude}. In our data, all of these 2,287 papers had
an overall decision of \texttt{exclude}. In addition, 2,240 had at least one exclude decision on the individual hard dealbreaker criteria. The table reports
unique-paper counts by each criterion. Note that a given paper can
contribute to multiple criteria (e.g., if it has both
registration/outcome-reporting problems and randomization or comparator
problems). 

The formal dealbreaker criteria are described in the rubric
(Table~\ref{tab:dealbreaker-framework}). As an example, ``Randomization'' does not mean that every paper in that row lacked an RCT label; rather, it means that Kurate
found the implemented assignment process to be absent, explicitly non-random or
quasi-random, unsupported by usable methods evidence, or compromised enough to
threaten the study's handling of potential selection biases. Common such issues included retroactive registration, evidence of outcome
switching, and unclear reporting of results from terminated or withdrawn trials.

\begin{table}[H]
\centering
\small
\setlength{\tabcolsep}{5pt}
\caption{A breakdown for the papers which failed on the individual dealbreaker criteria in our current sample, specifically the 2,287 papers with a final evidence-use decision of \texttt{exclude}.}
\label{tab:exclusion-breakdown}
\begin{tabular}{p{8.2cm}rr}
\toprule
Dealbreaker criterion & Papers & Excluded papers \\
\midrule
Control group & 124 & 5.4\% \\
Comparator validity & 444 & 19.4\% \\
Randomization & 712 & 31.1\% \\
Blinding and outcome susceptibility & 398 & 17.4\% \\
Analysis aligned with research question & 183 & 8.0\% \\
ITT or justified alternative & 320 & 14.0\% \\
Attrition and missing data & 102 & 4.5\% \\
Prospective registration and outcome specification & 952 & 41.6\% \\
Selective reporting and multiplicity & 1,358 & 59.4\% \\
\bottomrule
\end{tabular}
\end{table}
The dimensions on which papers exhibited issues were statistical
power (64.4\% of papers), selective reporting (63.8\%), and analysis
prespecification (60.0\%). This is somewhat unsurprising, given the extensive discourse around these issues in metascience. Infectious disease/HIV, neurology/pain, cardiometabolic, and cancer studies had the highest overall mean scores, whereas reproductive/pregnancy and mental
health studies had the lowest. Interestingly, the proportion of
exclude decisions did not follow the same ranking: for example,
gastrointestinal/IBD studies had a relatively low mean score, but a lower
exclusion rate than several higher-scoring domains. This reflects the fact that many of the dealbreaker issues did not necessarily map entirely onto the average score of these studies. 

Scores also increased over
time (Figure~\ref{fig:domain-trajectories}). The proportion of papers graded A,
B, or C rose from 4.3\% of papers published before 2000, and 20.1\% of those
published in 2000--2009, to 71.0\% of those published in 2020--2026. This
pattern may partly reflect broader availability of registrations, protocols, and
reporting infrastructure over time, not only improvements in underlying research
quality.

\subsection{Paper-level quality analysis}
Kurate's primary function, and its unique quality relative to many other automated research quality assessment tools, is its capacity to quickly and easily provide large-scale summaries of literatures and corpora. However, Kurate also produces a detailed assessment of each individual paper which can in principle be useful to researchers. To illustrate this, we now provide an example report derived from the SPRINT trial, a highly influential randomized trial which compared intensive and standard targets for systolic blood pressure in patients at high cardiovascular risk \citep{sprint2015}. Kurate assigns the SPRINT paper an A grade, with a normalized score of 0.9236. Table~\ref{tab:sprint} shows how Kurate evaluates the paper on each dimension, including by comparing the paper against the trial's public ClinicalTrials.gov registration \citep{sprintregistry2010}. Appendix Figure~\ref{fig:sprint-report} shows the summary at the top of Kurate's report for SPRINT.

\begingroup
\footnotesize
\setlength{\tabcolsep}{2.5pt}
\setlength{\LTleft}{\fill}
\setlength{\LTright}{\fill}
\renewcommand{\arraystretch}{1.06}
\begin{longtable}{@{}P{1.75cm}P{0.8cm}P{3.0cm}P{3.2cm}P{4.1cm}@{}}
  \caption{Kurate's assessment of the SPRINT trial on each scored dimension. Scores are normalized to lie between 0 and 1. The two middle columns show the evidence Kurate compared, with the source document given in square brackets: what was planned or expected (mainly from the registration or protocol), and what was reported.}
  \label{tab:sprint} \\
  \toprule
  \textbf{Dimension} & \textbf{Score} & \textbf{Planned or external evidence} & \textbf{Reported evidence} & \textbf{Kurate's assessment} \\
  \midrule
  \endfirsthead
  \toprule
  \textbf{Dimension} & \textbf{Score} & \textbf{Planned or external evidence} & \textbf{Reported evidence} & \textbf{Kurate's assessment} \\
  \midrule
  \endhead
  \midrule
  \multicolumn{5}{r}{\emph{Continued on next page}} \\
  \midrule
  \endfoot
  \bottomrule
  \endlastfoot
  Statistical power & 1.00 & [registry] planned enrollment ``9,250 participants'' & [report] ``88.7\% power \ldots 20\% effect''; ``randomly assigned 9361 persons'' & Strong support: the paper reports a specific power calculation, and the number of participants randomized exceeded the planned target. \\
  Causal identification & 0.67 & [registry] intensive vs standard systolic blood pressure target groups; comparison of two active treatments; limited detail on blinding & [report] ``We randomly assigned 9361 persons \ldots'' to ``intensive'' vs ``standard'' treatment & Partial support: the paper clearly reports random assignment, but evidence on allocation concealment (i.e., preventing those enrolling participants from knowing the upcoming assignment) and on blinding is not fully explicit in the paper. \\
  Preregistration & 1.00 & [registry] first submitted 2010-09-20, before study start on 2010-10-01; \texttt{NCT01206062} & [report] ``ClinicalTrials.gov number, NCT01206062''; the same trial is described in the article & Strong support: Kurate links the article to a registration which was submitted before the trial began, and verifies the dates. \\
  Selective reporting & 1.00 & [registry] primary outcome ``MI \ldots ACS \ldots Stroke \ldots HF \ldots CVD Death''; secondary outcome ``All-cause Mortality'' & [abstract] same composite primary outcome; ``All-cause mortality was also significantly lower \ldots'' & Strong support: the main reported outcomes match the preregistered outcomes, and the early stopping of the trial is reported openly. \\
  Measurement validity & 1.00 & [registry] primary outcome is a composite of major cardiovascular events & [report] ``primary composite outcome was myocardial infarction \ldots stroke \ldots heart failure \ldots'' & Strong support: the outcome is a standard clinical measure and is directly relevant to the question the trial addresses. \\
  Analysis prespecification & 1.00 & [protocol / registry] registered in advance; protocol available; planned ``time to the first occurrence \ldots'' analysis using a Cox model (a standard model for time-to-event data) on randomized participants & [results] ``stopped early \ldots owing to'' lower primary-event rates; the change from the plan is stated explicitly & Strong support: Kurate finds an analysis plan specified in advance, and the major change to that plan (stopping the trial early) is reported explicitly rather than made without disclosure. \\
  Reporting transparency & 0.83 & [Kurate's reporting criteria] participant flow, harms, and the set of participants analyzed should be reported in a form which can be checked & [Figure 1] ``Eligibility, Randomization, and Follow-up''; ``9361 participants were enrolled \ldots (Fig. 1)''; adverse events are reported & Adequate support: participant flow and the main outcomes can be checked, but Kurate was not able to fully extract how missing data were handled or handling of participant withdrawals. \\
  Conflict of interest and funding & 0.75 & [registry] ``Lead Sponsor: National Heart, Lung, and Blood Institute'' & [report] ``Funded by the National Institutes of Health''; ``Disclosure: forms \ldots available \ldots at NEJM.org'' & Partial support: public funding is clearly stated, but the paper does not report individual authors' conflicts of interest directly, and instead refers readers to separate disclosure forms. \\
\end{longtable}
\endgroup

\section{Limitations \& Conclusion}

Section~\ref{sec:benchmark} evaluates Kurate's information extraction against expert human labels. On the other hand, Kurate's scores, grades, and evidence-use decisions have yet to receive a similar evaluation. This is because no comparable expert-annotated benchmark currently exists for these judgments. In addition, we compared Kurate against the benchmark only for the subset of items which Kurate extracts, which may make the comparison with the human and baseline figures more favorable to Kurate than a comparison over the full benchmark would be. In practice, Kurate's outputs may contain model errors, failures to extract information, and errors in recording which document a piece of information came from. It is therefore critical that Kurate be extensively evaluated before formal deployment and use, ideally against a human rater panel which extracts and evaluates manuscripts along the full span of Kurate's quality evaluation dimensions. The use of a rater panel comparator is especially important for the cases where quality evaluation criteria are ambiguous even for human reviewers and reasonably scholarly disagreement means that there is no single ``correct'' answer. 

It is worth being explicit that the goal of Kurate, like other metascientific literature search tools, is to supplement human judgement and not to displace it. Indeed, beyond the principled reasons not to do so (i.e., because the final decision from scientific reasoning and evidence evaluation must be based on human judgement), there are also practical constraints: for instance the availability of related documents (e.g., protocols and registrations) varies substantially across journals and publication years, meaning that Kurate's records are necessarily incomplete. Further still, the evaluation decisions within Kurate (particularly the designation of hard dealbreaker criteria) inherently have a degree of subjectivity to them. Other researchers may disagree on, for example, the relative importance of registration-paper consistency (or how precisely consistency ought to be defined), among many other possible issues. As such, it is important to reiterate that Kurate is intended as a decision-support tool for researchers evaluating research quality, and not as a canonical or definitive judge on the matter. 

These limitations and considerations aside, the benchmark results show that the information Kurate extracts matched expert labels in 91\% of cases for trial protocols and 79\% of cases for results publications, which is comparable to (or better than) other published automated system. Although improvements are required in the numeric information extraction, it should be noted that this numeric information is currently not central to Kurate's evidence evaluation (i.e., it does not run numeric consistency checks such as GRIM or GRIMMER; \cite{brown2017grim, anaya2016grimmer}). We have also shown how Kurate can convert thousands of papers into a structured record of evidence quality, which supports appraisal at both the individual-study level, and (mostly important) at the level of a corpus.

\section{Author Comment}

When we assess scientific studies systematically against established standards for
research quality, substantial problems are found to be common, including in
randomized controlled trials, medical research, and research supported by major
public funders. Many of the individual problems we identify have been recognized
and discussed for some time. We believe that results like these deserve to be
communicated and interpreted carefully. Identifying weaknesses in the scientific
record is not the same as establishing what society should do in response to them,
and evidence of poor research quality can support very different conclusions about
funding, regulation, oversight, and public trust. We would not want work intended
to strengthen science to be interpreted, without that distinction, as a reason to
weaken the institutions that make science possible. We therefore think it is
important to be explicit about why we undertook this work, how our own experiences
have shaped our interpretation of the findings, and what we believe constructive
responses to them might look like.

\textbf{Matthew Vowels.}
My own perspective has changed considerably through moving from engineering and
machine learning into applications in psychology and medicine. I learned quickly
that research involving people presents specific challenges which are much less
common in the engineering sciences: effects can be small, experiments can be
expensive or even impossible to repeat, measurements can be noisy and indirect,
and the populations can be highly heterogeneous. This has made me more sympathetic
to the constraints under which empirical science operates, but also more concerned
about the problems that are within our control. A clinical trial may take years,
cost millions, expose patients to an intervention, and ultimately influence how
well people are treated. In that context, weaknesses in design, analysis, reporting,
and interpretation are not minor technicalities. I absolutely do not see these
findings as an argument for less science or less funding. I see them as an argument
for investing much more seriously in the infrastructure of good science:
methodological support, transparency, replication, independent scrutiny, and
scalable quality control. The purpose of Kurate is not to make it easier to distrust
science, but to make it easier to distinguish evidence that deserves our confidence
from evidence that does not.

\textbf{Jamie Cummins.}
As a metascientist, most of my tool-building work is focused on making it easier for people to do things which are healthy for the scientific literature but which are difficult to do (and therefore infrequently done). Kurate is one example of this: it aims to make hard-to-generate information about studies accessible, particularly with regard to formal quality checks in line with emerging standards (e.g., INSPECT-SR; \citealp{inspect-sr}). I see findings like those reported here as part of a healthy error culture in science: it is good for us to formally evaluate studies in these terms, and to discuss them openly and transparently. 

At the same time, I do not think that Kurate's evaluations should be seen as a canonical account of the quality of literature which everyone should endorse and use. Kurate is fallible in its judgements, and those judgements should not be treated as the definitive assessment of a study's quality. Although the dimensions we assess are based on established standards and emerging best practices, several of the calls involved (e.g., how evidence is weighted across dimensions) are ultimately subjective. I hope that readers primarily take away from this the proof-of-principle that large-scale quality assessment of this kind is feasible, together with a demonstration of what it can look like in practice. The specific evaluations provided by Kurate can and will be refined over time, not least through further validation of Kurate at a larger scale and against more elaborate human assessments of quality. 
\section{AI-Use Statement}
Large language models are part of the Kurate system itself, as described in Section~\ref{sec:kurate}, where they are used to extract information and in other stages of the evidence-appraisal pipeline. Separately from this use within the system, generative AI tools were used during the development of the software and the preparation of this manuscript. This included assistance with writing, debugging, restructuring, and documenting code; developing and revising parts of the pipeline; and drafting, editing, restructuring, paraphrasing, and proofreading parts of the manuscript.

\newpage

\bibliographystyle{plainnat}
\bibliography{neurips-workshop}

@book{cochranehandbook2023,
  editor = {Higgins, Julian P. T. and Thomas, James and Chandler, Jacqueline and Cumpston, Miranda and Li, Tianjing and Page, Matthew J. and Welch, Vivian A.},
  title = {Cochrane Handbook for Systematic Reviews of Interventions},
  publisher = {Cochrane},
  year = {2023},
  note = {Version 6.4, updated August 2023},
  url = {https://www.cochrane.org/handbook}
}

@article{alexander2021aducanumab,
  author  = {Alexander, G. Caleb and Emerson, Scott and Kesselheim, Aaron S.},
  title   = {Evaluation of Aducanumab for {Alzheimer} Disease: Scientific Evidence and Regulatory Review Involving Efficacy, Safety, and Futility},
  journal = {JAMA},
  year    = {2021},
  volume  = {325},
  number  = {17},
  pages   = {1717--1718},
  doi     = {10.1001/jama.2021.3854}
}

@article{grade2008,
  author = {Guyatt, Gordon H. and Oxman, Andrew D. and Vist, Gunn E. and Kunz, Regina and Falck-Ytter, Yngve and Alonso-Coello, Pablo and Sch{\"u}nemann, Holger J.},
  title = {{GRADE}: An Emerging Consensus on Rating Quality of Evidence and Strength of Recommendations},
  journal = {BMJ},
  year = {2008},
  volume = {336},
  number = {7650},
  pages = {924--926},
  doi = {10.1136/bmj.39489.470347.AD}
}

@article{borah2017analysis,
  author = {Borah, Rohit and Brown, Andrew W. and Capers, Patrice L. and Kaiser, Kathryn A.},
  title = {Analysis of the Time and Workers Needed to Conduct Systematic Reviews of Medical Interventions Using Data from the {PROSPERO} Registry},
  journal = {BMJ Open},
  year = {2017},
  volume = {7},
  number = {2},
  pages = {e012545},
  doi = {10.1136/bmjopen-2016-012545}
}

@article{anderson2015compliance,
  author = {Anderson, Monique L. and Chiswell, Karen and Peterson, Eric D. and Tasneem, Asba and Topping, Jane and Califf, Robert M.},
  title = {Compliance with Results Reporting at {ClinicalTrials.gov}},
  journal = {New England Journal of Medicine},
  year = {2015},
  volume = {372},
  number = {11},
  pages = {1031--1039},
  doi = {10.1056/NEJMsa1409364}
}

@article{mathieu2009comparison,
  author = {Mathieu, Sylvain and Boutron, Isabelle and Moher, David and Altman, Douglas G. and Ravaud, Philippe},
  title = {Comparison of Registered and Published Primary Outcomes in Randomized Controlled Trials},
  journal = {JAMA},
  year = {2009},
  volume = {302},
  number = {9},
  pages = {977--984},
  doi = {10.1001/jama.2009.1242}
}

@article{turner2008selective,
  author = {Turner, Erick H. and Matthews, Annette M. and Linardatos, Eftihia and Tell, Robert A. and Rosenthal, Robert},
  title = {Selective Publication of Antidepressant Trials and Its Influence on Apparent Efficacy},
  journal = {New England Journal of Medicine},
  year = {2008},
  volume = {358},
  number = {3},
  pages = {252--260},
  doi = {10.1056/NEJMsa065779}
}

@article{dechartres2013sample,
  author = {Dechartres, Agn{\`e}s and Trinquart, Ludovic and Boutron, Isabelle and Ravaud, Philippe},
  title = {Influence of Trial Sample Size on Treatment Effect Estimates: Meta-Epidemiological Study},
  journal = {BMJ},
  year = {2013},
  volume = {346},
  pages = {f2304},
  doi = {10.1136/bmj.f2304}
}

@article{prasad2015surrogate,
  author = {Prasad, Vinay and Kim, Chul and Burotto, Mauricio and Vandross, Andrae},
  title = {The Strength of Association Between Surrogate End Points and Survival in Oncology: A Systematic Review of Trial-Level Meta-Analyses},
  journal = {JAMA Internal Medicine},
  year = {2015},
  volume = {175},
  number = {8},
  pages = {1389--1398},
  doi = {10.1001/jamainternmed.2015.2829}
}

@article{sprint2015,
  author = {{SPRINT Research Group}},
  title = {A Randomized Trial of Intensive versus Standard Blood-Pressure Control},
  journal = {New England Journal of Medicine},
  year = {2015},
  volume = {373},
  number = {22},
  pages = {2103--2116},
  doi = {10.1056/NEJMoa1511939},
  pmid = {26551272},
  pmcid = {PMC4689591}
}

@misc{sprintregistry2010,
  author = {{ClinicalTrials.gov}},
  title = {Record History, Version 3: {NCT01206062} -- Systolic Blood Pressure Intervention Trial},
  year = {2010},
  url = {https://clinicaltrials.gov/study/NCT01206062?a=3&tab=history},
  note = {First submitted September 20, 2010; first posted September 21, 2010; version 3 dated December 6, 2010; accessed August 11, 2026}
}

@article{robotreviewer2016,
  author = {Marshall, Iain J. and Kuiper, Jo{\"e}l and Wallace, Byron C.},
  title = {RobotReviewer: Evaluation of a System for Automatically Assessing Bias in Clinical Trials},
  journal = {Journal of the American Medical Informatics Association},
  year = {2016},
  volume = {23},
  number = {1},
  pages = {193--201},
  doi = {10.1093/jamia/ocv044}
}

@article{trialstreamer2020,
  author = {Marshall, Iain J. and Nye, Benjamin and Kuiper, Jo{\"e}l and Noel-Storr, Anna and Marshall, Rachel and Maclean, Rory and Soboczenski, Frank and Nenkova, Ani and Thomas, James and Wallace, Byron C.},
  title = {Trialstreamer: A Living, Automatically Updated Database of Clinical Trial Reports},
  journal = {Journal of the American Medical Informatics Association},
  year = {2020},
  volume = {27},
  number = {12},
  pages = {1903--1912},
  doi = {10.1093/jamia/ocaa163}
}

@misc{regcheck2026,
  author = {Cummins, Jamie and Clarke, Beth and Hussey, Ian and Elson, Malte},
  title = {{RegCheck}: A Tool for Structured Comparisons Between Study Registrations and Papers},
  year = {2026},
  eprint = {2601.13330},
  archivePrefix = {arXiv},
  primaryClass = {cs.CL},
  doi = {10.48550/arXiv.2601.13330},
  url = {https://arxiv.org/abs/2601.13330}
}

@article{compare2019,
  author = {Goldacre, Ben and Drysdale, Henry and Dale, Aaron and Milosevic, Ioan and Slade, Eirion and Hartley, Philip and Marston, Cicely and Powell-Smith, Anna and Heneghan, Carl and Mahtani, Kamal R.},
  title = {{COMPare}: A Prospective Cohort Study Correcting and Monitoring 58 Misreported Trials in Real Time},
  journal = {Trials},
  year = {2019},
  volume = {20},
  number = {1},
  pages = {118},
  doi = {10.1186/s13063-019-3173-2},
  pmid = {30760329},
  pmcid = {PMC6375128}
}

@article{menke2020sciscore,
  author  = {Menke, Joe and Roelandse, Martijn and Ozyurt, Burak and Martone, Maryann and Bandrowski, Anita},
  title   = {The Rigor and Transparency Index Quality Metric for Assessing Biological and Medical Science Methods},
  journal = {iScience},
  year    = {2020},
  volume  = {23},
  number  = {11},
  pages   = {101698},
  doi     = {10.1016/j.isci.2020.101698},
  pmcid   = {PMC7644557}
}

@misc{reviewerzero2026,
  author       = {{ReviewerZero AI, Inc.}},
  title        = {{ReviewerZero}: {AI}-Powered Research Integrity Platform},
  year         = {2026},
  howpublished = {\url{https://www.reviewerzero.ai}},
  note         = {Accessed October 4, 2026}
}

@article{anaya2016grimmer,
  author  = {Anaya, Jordan},
  title   = {The {GRIMMER} Test: A Method for Testing the Validity of Reported Measures of Variability},
  journal = {PeerJ Preprints},
  year    = {2016},
  volume  = {4},
  pages   = {e2400v1},
  doi     = {10.7287/peerj.preprints.2400v1},
  note    = {Preprint}
}

@article{syed2025,
  author = {Syed, Moin},
  title = {Some Data Indicating That Editors and Reviewers Do Not Check Preregistrations During the Review Process},
  journal = {Journal of Trial and Error},
  year = {2025},
  volume = {6},
  number = {1},
  doi = {10.36850/e5ce-4cc5},
  url = {https://journal.trialanderror.org/pub/check-preregistrations}
}

@article{nuijten2016statcheck,
  author = {Nuijten, Mich{\`e}le B. and Hartgerink, Chris H. J. and van Assen, Marcel A. L. M. and Epskamp, Sacha and Wicherts, Jelte M.},
  title = {The Prevalence of Statistical Reporting Errors in Psychology (1985--2013)},
  journal = {Behavior Research Methods},
  year = {2016},
  volume = {48},
  number = {4},
  pages = {1205--1226},
  doi = {10.3758/s13428-015-0664-2},
  pmid = {26497820},
  pmcid = {PMC5101263}
}

@article{brown2017grim,
  author = {Brown, Nicholas J. L. and Heathers, James A. J.},
  title = {The {GRIM} Test: A Simple Technique Detects Numerous Anomalies in the Reporting of Results in Psychology},
  journal = {Social Psychological and Personality Science},
  year = {2017},
  volume = {8},
  number = {4},
  pages = {363--369},
  doi = {10.1177/1948550616673876}
}

@manual{jung2025scrutiny,
  author = {Jung, Lukas},
  title = {{scrutiny}: Error Detection in Science},
  year = {2025},
  note = {R package version 0.6.1},
  url = {https://github.com/lhdjung/scrutiny},
  doi = {10.32614/CRAN.package.scrutiny}
}

@manual{metacheck2026,
  author = {DeBruine, Lisa and Mesquida, Cristian and Werner, Jakub and Lakens, Daniel},
  title = {{metacheck}: Check Research Outputs for Best Practices},
  year = {2026},
  note = {R package version 0.1.0},
  url = {https://scienceverse.org/metacheck},
  doi = {10.5281/zenodo.20704754}
}

@article{inspect-sr,
  author = {Wilkinson, Jack D. and Heal, Calvin and Flemyng, Ella and Antoniou, Georgios A. and Aburrow, Tony and Alfirevic, Zarko and Avenell, Alison and others},
  title = {{INSPECT-SR}: A Tool for Assessing Trustworthiness of Randomised Controlled Trials},
  journal = {medRxiv},
  year = {2025},
  doi = {10.1101/2025.09.03.25334905},
  note = {Preprint},
  url = {https://www.medrxiv.org/content/10.1101/2025.09.03.25334905}
}

@inproceedings{inspect-ai,
  author = {Markovic, Milan and Indukuri, Goutham and Sripada, Somayajulu and Vorland, Colby J. and Wilkinson, Jack and Robertson, Clare and Bolland, Mark J. and Grey, Andrew and Brazzelli, Miriam and Avenell, Alison},
  title = {Authoring and Management of Transparent Research Integrity Assessments of Randomised Clinical Trial Publications Using {LLM}-assisted Tools and Provenance Knowledge Graphs},
  booktitle = {International Semantic Web Conference},
  series = {Lecture Notes in Computer Science},
  publisher = {Springer Nature},
  year = {2026},
  note = {Accepted/In press},
  url = {https://abdn.elsevierpure.com/en/publications/authoring-and-management-of-transparent-research-integrity-assess/}
}

@article{spiritconsortelm,
  author = {Jiang, Lan and Ying, Xiangci and Brown, Andrew W. and Lan, Mengfei and Song, Wenxin and Menke, Joe D. and Vorland, Colby J. and Mayo-Wilson, Evan and Kilicoglu, Halil},
  title = {{SPIRIT-CONSORT-ELM}: Element-Level Annotated Dataset and Large Language Model Approach for Assessing Randomized Controlled Trial Reporting},
  journal = {medRxiv},
  year = {2026},
  pages = {2026.06.06.26354746},
  doi = {10.64898/2026.06.06.26354746},
  pmid = {42369465},
  pmcid = {PMC13308290},
  note = {Preprint},
  url = {https://pmc.ncbi.nlm.nih.gov/articles/PMC13308290/}
}

@article{spirit2013,
  author = {Chan, An-Wen and Tetzlaff, Jennifer M. and Altman, Douglas G. and Laupacis, Andreas and G{\o}tzsche, Peter C. and Krle{\v{z}}a-Jeri{\'c}, Karmela and Hr{\'o}bjartsson, Asbj{\o}rn and Mann, Howard and Dickersin, Kay and Berlin, Jesse A. and Do{\'r}e, Caroline J. and Parulekar, Wendy R. and Summerskill, William S. M. and Groves, Trish and Schulz, Kenneth F. and Sox, Harold C. and Rockhold, Frank W. and Rennie, Drummond and Moher, David},
  title = {{SPIRIT} 2013 Statement: Defining Standard Protocol Items for Clinical Trials},
  journal = {Annals of Internal Medicine},
  year = {2013},
  volume = {158},
  number = {3},
  pages = {200--207},
  doi = {10.7326/0003-4819-158-3-201302050-00583}
}

@article{consort2010,
  author = {Schulz, Kenneth F. and Altman, Douglas G. and Moher, David},
  title = {{CONSORT} 2010 Statement: Updated Guidelines for Reporting Parallel Group Randomised Trials},
  journal = {BMJ},
  year = {2010},
  volume = {340},
  pages = {c332},
  doi = {10.1136/bmj.c332}
}

@article{gwet2008,
  author = {Gwet, Kilem Li},
  title = {Computing Inter-rater Reliability and Its Variance in the Presence of High Agreement},
  journal = {British Journal of Mathematical and Statistical Psychology},
  year = {2008},
  volume = {61},
  number = {1},
  pages = {29--48},
  doi = {10.1348/000711006X126600}
}

@article{wilson1927,
  author = {Wilson, Edwin B.},
  title = {Probable Inference, the Law of Succession, and Statistical Inference},
  journal = {Journal of the American Statistical Association},
  year = {1927},
  volume = {22},
  number = {158},
  pages = {209--212},
  doi = {10.1080/01621459.1927.10502953}
}
\clearpage
\appendix

\section{Benchmark methodology and detailed results}
\label{sec:benchmark-appendix}

SPIRIT-CONSORT-ELM \elmcite{} consists of 200 documents: 100 clinical-trial
protocols, each paired with the publication which reports that trial's results.
For each item of the relevant reporting checklist, human annotators recorded the
exact text associated with that item, and these items are operationalized as 119
questions which each target a specific piece of reported information. The
protocol questions are based on SPIRIT items \citep{spirit2013}, which concern
what should be reported about a trial before it begins. The results questions,
on the other hand, are based on CONSORT items \citep{consort2010}, which concern
what should be reported once a trial has been completed.

The benchmark provides its own division of documents into training, test, and
validation sets, which we also used. All development work used the training set
(70 documents of each type). The benchmark's official test and validation sets,
combined (30 documents of each type), were held out and scored once for each
document type. All 100 results publications were already present in the corpus
described in Appendix~\ref{sec:corpus-targeting}. The 100 protocols, by contrast,
were collected specifically for this evaluation and are not part of that corpus.

\subsection{Two accuracy measures}

We mapped each benchmark item onto the corresponding structured field in Kurate,
and scored two measures. The first measure, \textbf{candidate extraction}, is
whether the correct value appears anywhere among the candidate values which
Kurate extracted for that field, and indicates whether Kurate found the
information at all. The second measure, \textbf{final value}, is whether the
single value which Kurate passes on to its later stages (i.e., scoring, the
evidence-use decision, and reporting) is correct; this is the measure reported
in the main text. We also report the number of scored items, $n$, since this
determines the precision of each estimate (i.e., the width of its confidence
interval). We score each extraction target only in documents where the benchmark includes
a corresponding ground-truth annotation. As a result, denominators differ between targets.

\begin{table}[h]
\centering
\small
\setlength{\tabcolsep}{4pt}
\caption{Full results for the training and held-out sets. The held-out columns use the benchmark's own test and validation sets combined, scored once for each document type. $k$ is the number of benchmark questions for which Kurate has a corresponding field. AC1 is the mean ($\pm$ SD) of the coefficients calculated separately for each question, which matches the method used by \elmcite{}.}
\label{tab:benchmark-full}
\begin{tabular}{lcccc}
\toprule
& \multicolumn{2}{c}{Protocols (SPIRIT)} & \multicolumn{2}{c}{Results (CONSORT)} \\
\cmidrule(lr){2-3}\cmidrule(lr){4-5}
& Train & Held-out & Train & Held-out \\
\midrule
Documents & 70 & 30 & 70 & 30 \\
Scored items $n$ & 566 & 242 & 906 & 370 \\
Questions $k$ & 9 & 9 & 15 & 15 \\
\midrule
Candidate extraction & 0.921 & \textbf{0.934} & 0.846 & \textbf{0.838} \\
\quad 95\% CI \citep{wilson1927} & --- & [0.895, 0.959] & --- & [0.797, 0.872] \\
Final value & 0.919 & 0.913 & 0.802 & 0.795 \\
Mean AC1 & --- & $0.937 \pm 0.103$ & --- & $0.813 \pm 0.173$ \\
\quad 95\% CI of the mean & --- & [0.870, 1.000] & --- & [0.725, 0.900] \\
F1 & --- & 0.934 & --- & 0.879 \\
\midrule
\multicolumn{5}{l}{\textit{Reference figures} \elmcite\textit{, each a mean over all 119 questions}} \\
\quad Human vs.\ consensus labels & \multicolumn{4}{c}{AC1 $0.891$, $0.893$} \\
\quad Baseline vs.\ consensus labels & \multicolumn{4}{c}{AC1 $0.796$ / F1 $0.822$} \\
\quad Human inter-annotator & \multicolumn{4}{c}{AC1 $0.782 \pm 0.219$} \\
\bottomrule
\end{tabular}
\end{table}

Table~\ref{tab:benchmark-full} reports the results for both sets. For both
document types and both measures, performance on the held-out documents was
within 1.3 percentage points of performance on the training documents (e.g.,
candidate extraction on protocols was $0.934$ on held-out documents and $0.921$
on training documents), indicating little evidence of overfitting. 

The benchmark's authors publish three reference figures \elmcite{}. The
agreement of each individual annotator with the final consensus labels (AC1
$0.891$ and $0.893$) is the figure most comparable to ours, because Kurate is
also scored against those consensus labels. The agreement between the two human
annotators (AC1 $0.782$) instead measures how much two experts agree with each
other. The third figure is the agreement of the baseline system with the
consensus labels (AC1 $0.796$, F1 $0.822$).

\subsection{Results for each benchmark item}

Tables~\ref{tab:benchmark-items-protocol} and
\ref{tab:benchmark-items-results} give the counts for each item under the
benchmark's own item identifiers, for both measures. These counts are combined
to produce the figures in Table~\ref{tab:benchmark}.

\begin{table}[h]
\centering
\footnotesize
\setlength{\tabcolsep}{4pt}
\caption{Protocols, held-out set ($n=242$).}
\label{tab:benchmark-items-protocol}
\begin{tabular}{@{}>{\raggedright\arraybackslash}p{5.0cm}>{\raggedright\arraybackslash}p{3.9cm}cc@{}}
\toprule
Benchmark item & Kurate field & Candidate extraction & Final value \\
\midrule
\texttt{text10a\_participants\_inclusion} & eligibility criteria & 30/30 & 30/30 \\
\texttt{text11a\_intervention\_description} & intervention name & 30/30 & 30/30 \\
\texttt{text11a\_intervention\_description} & comparator name & 30/30 & 30/30 \\
\texttt{text14b\_sample\_calculation} & sample size & 30/30 & 30/30 \\
\texttt{text20a\_statistical\_methods\_outc} & planned analysis methods & 29/29 & 29/29 \\
\texttt{text31d\_sharing\_data} & data-sharing statement & 5/5 & 5/5 \\
\texttt{text14b\_sample\_calculation} & location of power calculation & 27/30 & 26/30 \\
\texttt{text12a\_outcomes\_definitions} & endpoint name & 24/30 & 24/30 \\
\texttt{text4\_funding} & funding & 21/28 & 17/28 \\
\midrule
\textbf{Total} & & \textbf{226/242} & \textbf{221/242} \\
\bottomrule
\end{tabular}
\end{table}

\begin{table}[h]
\centering
\footnotesize
\setlength{\tabcolsep}{4pt}
\caption{Results publications, held-out set ($n=370$).}
\label{tab:benchmark-items-results}
\begin{tabular}{@{}>{\raggedright\arraybackslash}p{5.0cm}>{\raggedright\arraybackslash}p{3.9cm}cc@{}}
\toprule
Benchmark item & Kurate field & Candidate extraction & Final value \\
\midrule
\texttt{text14b\_sample\_calculation} & sample size & 23/23 & 20/23 \\
\texttt{text37a\_analysis\_numbers} & sample size & 23/23 & 18/23 \\
\texttt{text38a\_outcome\_results} & $p$-value & 22/22 & 21/22 \\
\texttt{text11a\_intervention\_description} & intervention name & 29/30 & 29/30 \\
\texttt{text38a\_outcome\_results} & summary statistic & 28/30 & 23/30 \\
\texttt{text12a\_outcomes\_definitions} & endpoint name & 26/29 & 26/29 \\
\texttt{text38b\_binary\_results} & summary statistic & 14/16 & 12/16 \\
\texttt{text11a\_intervention\_description} & comparator name & 25/30 & 25/30 \\
\texttt{text38a\_outcome\_results} & effect size & 25/30 & 25/30 \\
\texttt{text31d\_sharing\_data} & data-sharing statement & 15/18 & 15/18 \\
\texttt{text4\_funding} & funding & 22/27 & 22/27 \\
\texttt{text38b\_binary\_results} & effect size & 12/16 & 12/16 \\
\texttt{text38a\_outcome\_results} & confidence interval & 19/30 & 19/30 \\
\texttt{text38b\_binary\_results} & prevalence / percentage & 10/16 & 10/16 \\
\texttt{text14b\_sample\_calculation} & location of power calculation & 17/30 & 17/30 \\
\midrule
\textbf{Total} & & \textbf{310/370} & \textbf{294/370} \\
\bottomrule
\end{tabular}
\end{table}

\subsection{Limitations of the comparison}

First, the benchmark contains 119 questions, of which Kurate has structured
fields for 9 on protocols and 15 on results publications, whereas the published
human and baseline means are calculated over all 119 questions. The questions
which Kurate does not cover are plausibly harder on average than those it does
cover. Our means are therefore not directly comparable to the reference means,
and should be interpreted as performance on the subset of questions which Kurate
covers. A like-for-like comparison would require extending Kurate's fields to
cover the full set of questions.

Second, the protocols were collected specifically for this evaluation, and are
not part of the corpus analyzed in Section~\ref{sec:results}. Performance on
protocols therefore indicates how well Kurate extracts information from
documents written before a trial begins, but does not describe the composition
of that corpus.

\section{SPRINT Report}

\begin{figure}[H]
    \centering
    \makebox[\linewidth][c]{\includegraphics[width=1.3\linewidth]{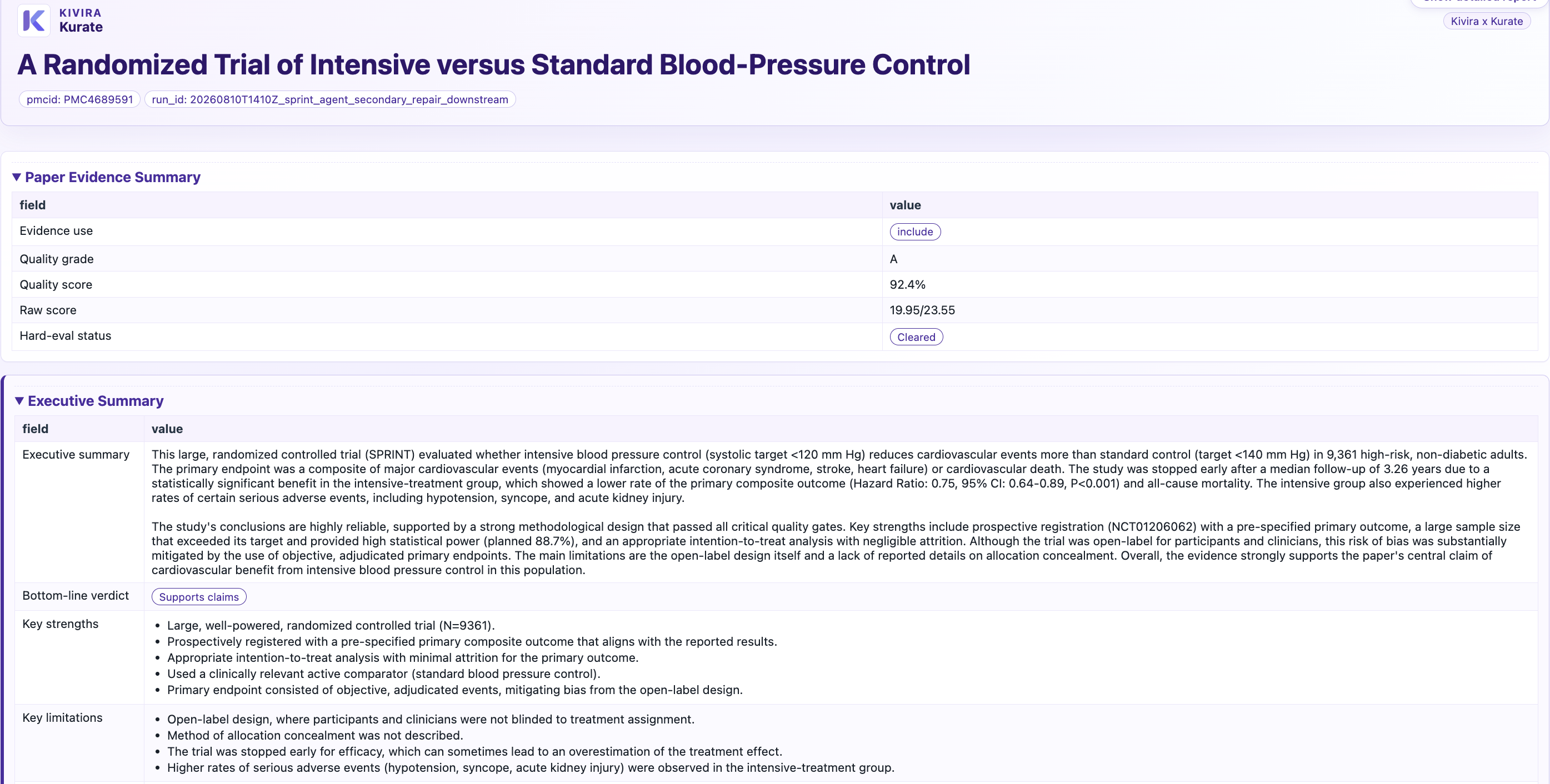}}
    \caption{Screenshot of the summary section at the top of Kurate's report for the SPRINT trial.}
    \label{fig:sprint-report}
\end{figure}

\section{Rubric}
\label{sec:rubric}
The Kurate rubric separates the overall quality grade from the evidence-use include/exclude decision. Firstly, it
assigns a quality grade by scoring several dimensions of study design and
reporting. During scoring, each dimension can either count against the
paper, or count in its favour, or be neutral: most dimensions range from $-3$ to
$3$, while conflict of interest and funding ranges from $-2$ to $2$. Negative
scores reflect flaws or missing required evidence, positive scores
reflect strengths, and zero means that the presence, absence, or nature of a certain feature is neutral.
Second, Kurate separately checks for hard eligibility / `dealbreaker' problems. These can limit
whether a paper is used as supporting evidence even if its overall numeric grade is not especially low.

\subsection{Scored quality dimensions}

\begin{table}[H]
  \centering
  \caption{Scored quality dimensions used to assign the Kurate grade.}
  \label{tab:scoring-policy}
  \small
  \setlength{\tabcolsep}{4pt}
  \begin{tabular}{P{2.9cm}P{6.3cm}cc}
    \toprule
    \textbf{Dimension} & \textbf{Short description} & \textbf{Range} & \textbf{Weight} \\
    \midrule
    Statistical power & Whether the study was large enough for its main claim, given the planned and analyzed sample sizes. & $[-3,3]$ & 1.35 \\
    Causal identification & Whether the design supports a causal comparison, including randomization, allocation safeguards, comparator choice, and blinding where feasible. & $[-3,3]$ & 1.35 \\
    Preregistration & Whether the main claim was specified before the relevant data were collected or analyzed. & $[-3,3]$ & 1.20 \\
    Selective reporting & Whether the reported outcomes, analyses, and emphasis match the plan made in advance. & $[-3,3]$ & 1.20 \\
    Measurement validity & Whether the outcomes are appropriate for the construct, population, intervention, and timepoint. & $[-3,3]$ & 1.00 \\
    Analysis prespecification & Whether the main analysis was planned in advance and answers the stated research question. & $[-3,3]$ & 1.00 \\
    Reporting transparency & Whether the paper reports enough methods, participant-flow, missing-data, safety, and source information for the study to be checked. & $[-3,3]$ & 0.45 \\
    Conflict of interest / funding & Whether funding, sponsor role, author conflicts, and related integrity signals affect interpretation. & $[-2,2]$ & 0.45 \\
    \bottomrule
  \end{tabular}
\end{table}

Before aggregation, each score is clamped to its configured range. Let
$\tilde{s}_i$ be that clamped score, $w_i$ the dimension weight, and
$[m_i,M_i]$ the configured range. Kurate computes
\[
  T=\sum_i w_i\tilde{s}_i,\qquad
  q=\min\{1,\max\{0,(T-\sum_i w_i m_i)/(\sum_i w_i M_i-\sum_i w_i m_i)\}\}.
\]
The normalized score $q$ is rounded to four decimals and mapped to a letter grade (A--F) using Table \ref{tab:grade-mapping}. If no active dimension has been scored, the grade is F. Where an expected dimension rating is absent, it is treated as neutral (i.e., it neither raises nor lowers $q$). However, the grade is marked as incomplete/provisional in this case. Runtime fallback scores are treated in the same way. Kurate also tracks the reliability of its own extraction by recording
extraction-reliability diagnostics, including whether core paper sections were recovered, whether parser or section-classification anomalies were raised, and how much of the evidence required for each rating was available. This facilitates further improvements of the workflow by identifying points of failure which can be improved in further development. Because we expect Kurate extractions to be imperfect (and indeed, our evaluations suggest this is certainty the case for numeric information), non-recovered values are not a priori treated as true absences. In particular, when other signals indicate the possibility that Kurate has failed to successfully extract true values, absence-based penalties which would otherwise affect the score, are `neutralised' or at least attenuated (depending on the presence of other evidence) and the absence is flagged for the user. In other words, absence is scored negatively only when there appears to be sufficient extraction success on other dimensions/elements that the absence is likely to indicate an actual absence and not a parsing failure.

\begin{table}[H]
  \centering
  \caption{Summary of Kurate's scoring and grade mapping.}
  \label{tab:grade-mapping}
  \small
  \setlength{\tabcolsep}{4pt}
  \begin{tabular}{P{3.1cm}P{4.5cm}P{5.0cm}}
    \toprule
    \textbf{Scoring element} & \textbf{Mapping} & \textbf{Interpretation} \\
    \midrule
    Dimension score & $-3$ to $+3$ for most active dimensions, with conflict/funding using a narrower range. & Negative scores are penalties for fundamental flaws or missing  evidence, with zero as neutral and positive scores indicating support. \\
    Evidence-grade score & Weighted dimension total normalized to 0--1. & Produces the continuous normalized score reported for papers and cohorts. \\
    A & normalized score $\geq 0.8333$ & Strong evidence quality across the active dimensions. \\
    B & normalized score $\geq 0.6667$ and $<0.8333$ & Adequate-to-strong evidence quality with limited concerns. \\
    C & normalized score $\geq 0.5000$ and $<0.6667$ & Moderate evidence quality with notable limitations. \\
    D & normalized score $\geq 0.3333$ and $<0.5000$ & Limited evidence quality with substantial weaknesses. \\
    E & normalized score $\geq 0.1667$ and $<0.3333$ & Very weak evidence quality, but not in the lowest band. \\
    F & normalized score $<0.1667$ & Lowest configured quality band or no usable graded evidence. \\
    Dealbreaker eligibility & Separate include, flag, exclude, or processing-hold decision. & Confirmed dealbreaker failures likely render a paper unusable for evidence regardless of its grade. Borderline cases are flagged for manual review. \\
    \bottomrule
  \end{tabular}
\end{table}

\newpage
\subsection{Hard eligibility criteria (dealbreakers)}

The evidence-use state (i.e., include or exclude) is assigned separately from the grade/scoring. Once quality/dimension
scoring is completed, the evidence-use defaults to \texttt{include}; at this point, grade thresholds are retained for
governance and for when the user submits a query, rather than used as automatic exclusion
gates. For example, later one, once the DB is live and being interacted with by users,a user could set their standard of evidence threshold to be grade `C' meaning all studies with a grade lower than this are excluded from consideration when evaluating a user query. Given the stored intermediate outputs, the final
\texttt{include}/\texttt{flag}/\texttt{exclude} assignment is deterministic: the system applies fixed rules to the
completed score profile, hard-evaluation annotation, runtime status, and
top-level reliability checks. It is worth noting that the hard-evaluation and executive-assessment annotations themselves are LLM-assisted judgments which are assigned during the study processing (i.e., upstream of a user's query/filtering). We have not
yet run an evaluation of the system's judgements under repeated-runs, which would help to understand the stability and consistency of Kurate.
Importantly, the dealbreaker-evaluation can override the default `include' state, so that any serious flaws are not simply averaged away by otherwise high 
scores. Table~\ref{tab:scoring-policy} summarises the scored quality dimensions, and Table~\ref{tab:dealbreaker-framework} summarises the hard eligibility criteria. Table~\ref{tab:exclusion-breakdown} in
Section~\ref{sec:results} reports how often those hard criteria drove
exclusions in the current snapshot.

\begin{table}[H]
\centering
\small
\setlength{\tabcolsep}{4pt}
\renewcommand{\arraystretch}{1.08}
\caption{Hard dealbreaker criteria that can flag or exclude a paper from evidence use.}
\label{tab:dealbreaker-framework}
\begin{tabular}{P{4.0cm}P{9.0cm}}
\toprule
\textbf{Hard dealbreaker} & \textbf{Short description} \\
\midrule
Control group & The study needs a usable comparison group for the claim being made. A causal claim cannot rest on a paper with no suitable control or comparison. \\
Comparator validity & The comparator must be clear and appropriate. A very weak, poorly described, or inappropriate comparator can make the claimed effect uninterpretable. \\
Randomization & Assignment should be random, or at least not presented as randomized when it was absent, quasi-random, unsupported, or compromised in a way that could bias treatment assignment. \\
Blinding and outcome susceptibility & Lack of blinding is most serious when the primary outcome is subjective or otherwise easy to bias. Lack of blinding is not automatically fatal when blinding is infeasible and the claim is limited accordingly. \\
Analysis aligned with research question & The model, comparator, outcome, timepoint, and analysis population should answer the stated research question. If they do not, the primary claim may not be supported. \\
ITT or justified alternative & The analysis should preserve the randomized comparison, usually by intention-to-treat (ITT), or give a credible reason for an alternative population. Unjustified completer-only, subgroup, or as-treated analyses can undermine the comparison. \\
Attrition and missing data & Dropout and missing data should not be severe, highly imbalanced between groups, or handled in a way that changes the meaning of the result. \\
Prospective registration and outcome specification & The main claim should be traceable to a credible registration, protocol, or analysis plan made before the relevant data were collected, analyzed, or reported. Late, current-only, or unverified registration evidence is treated cautiously. \\
Selective reporting and multiplicity & Reported outcomes and analyses should match what was specified in advance. Outcome switching, omitted registered outcomes, unexplained subgroup emphasis, post-hoc confirmatory claims, or unhandled multiple comparisons can restrict evidence use. \\
\bottomrule
\end{tabular}
\end{table}

\clearpage

\end{document}